# Evaluating Anytime Algorithms for Learning Optimal Bayesian Networks

**Brandon Malone**
Department of Computer Science
Helsinki Institute for Information Technology
Fin-00014 University of Helsinki, Finland
brandon.malone@cs.helsinki.fi

**Changhe Yuan**
Department of Computer Science
Queens College/City University of New York
Queens, NY 11367 USA
changhe.yuan@qc.cuny.edu

## Abstract

Exact algorithms for learning Bayesian networks guarantee to find provably optimal networks. However, they may fail in difficult learning tasks due to limited time or memory. In this research we adapt several anytime heuristic search-based algorithms to learn Bayesian networks. These algorithms find high-quality solutions quickly, and continually improve the incumbent solution or prove its optimality before resources are exhausted. Empirical results show that the anytime window A* algorithm usually finds higher-quality, often optimal, networks more quickly than other approaches. The results also show that, surprisingly, while generating networks with few parents per variable are structurally simpler, they are harder to learn than complex generating networks with more parents per variable.

## 1 INTRODUCTION

Score-based learning of Bayesian networks is a popular strategy which assigns a score to a network structure based on given data, and the goal is to find the highest-scoring structure. The problem is NP-complete (Chickering 1996), so much early research focused on local search strategies, such as greedy hill climbing in the space of Bayesian network structures (Heckerman 1996), hill climbing in the space of equivalence classes of networks (Chickering 2002) or hill climbing in the space of variable orderings (Teyssier and Koller 2005). Other more sophisticated local search techniques have also been investigated (Moore and Wong 2003). Unfortunately, these algorithms offer no bounds on the quality of learned networks. On the other hand, they do have good *anytime* behavior. That is, they quickly find a solution and improve its quality throughout the search. The search can be stopped at "anytime" and return the best solution found so far.

Several dynamic programming (DP) algorithms (Koivisto and Sood 2004; Ott, Imoto, and Miyano 2004; Singh and Moore 2005; Silander and Myllymaki 2006; Malone, Yuan, and Hansen 2011) have been developed which guarantee to find the highest scoring network for a dataset. However, these algorithms do not exhibit anytime behavior; they do not produce any solution until giving the optimal network at the end of the search.

Recently, though, several algorithms have been developed which include both optimality guarantees and anytime behavior. de Campos and Ji (2011) proposed a branch and bound algorithm (BB). It begins with a (cyclic) structure in which all variables have their optimal parents. Then, cycles are broken in a best-first manner until the optimal structure is found. These cyclic structures give a lower bound on the optimal network which improves throughout the search. To add anytime behavior, a local search algorithm is used to learn a suboptimal network at the beginning of the search. The score of that network serves as an upper bound. Furthermore, the search sometimes deviates from a pure best-first strategy to find acyclic structures and improve the upper bound. At anytime, the search can be stopped, and the ratio between the upper and lower bounds give a quality guarantee of the current best acyclic network. When the two bounds agree, the current best structure is optimal.

Mathematical programming (MP) algorithms (Jaakkola et al. 2010; Cussens 2011) have also been developed which have both anytime behavior and optimality guarantees. These algorithms search in a space which includes an embedded polytope whose surface corresponds to Bayesian networks. The polytope has exponentially many facets, so it is not represented explicitly. Rather, a series of MPs are solved to define the polytope and find the optimal point on its surface, which corresponds to the optimal Bayesian network. The points on the surface correspond to integer coordinates, so the MPs are actually integer linear programs (ILPs) which are solved by relaxing the problem to a normal linear program (LP). After solving each LP, the solution is checked for integrality. If it is integral, then the solution corresponds to the optimal Bayesian network. If not, the value of the solution gives a lower bound on the opti-mal score. Also, the solution can be used to decode a valid acyclic network and find an upper bound on the score. As with BB, at any time, the search can be stopped, and the ratio between the bounds gives a quality guarantee.

Yuan *et al.* (2011) described a shortest path formulation for the structure learning problem. Since then, several heuristic search algorithms, including A* (Yuan, Malone, and Wu 2011) and BFBnB (Malone et al. 2011), have been applied to this problem. This paper explores the empirical behavior of a variety of *anytime heuristic search algorithms*, including anytime weighted A* (AWeiA*) (Hansen and Zhou 2007), anytime repairing A* (ARA*) (Likhachev, Gordon, and Thrun 2003) and anytime window A* (AWinA*) (Aine, Chakrabarti, and Kumar 2007), within this shortest path formulation. Like BB and MP, these algorithms all incorporate optimality guarantees and anytime behavior. We empirically compare these algorithms and MP on a variety of synthetic datasets. We use synthetic datasets because these allow us to better control experimental conditions which affect the learning, including the number of variables, number of records and complexity of the generating process of the datasets.

Experimentally, we show that AWinA* outperforms the other anytime heuristic search algorithms. It also often finds higher-quality networks more quickly than MP, but is slower to prove optimality for simpler synthetic networks. More thorough investigation into the search space and runtime characteristics of the algorithms provide additional insight to the learning problem. In particular, our results show that complex generating networks may seem structurally challenging to learn, but they actually lie within or close to the promising solution space that is first explored by heuristic search and are thus easier to find. In contrast, simple generating networks typically receive bad estimated scores. Because many other search nodes have better score estimates, heuristic search cannot easily prove optimality for these datasets.

## 2 LEARNING BAYESIAN NETWORKS

A Bayesian network consists of a directed acyclic graph (DAG) in which vertices correspond to a set of random variables $V = \{X_1, ..., X_n\}$ and a set of conditional probability distributions. The arcs in the DAG encode conditional independence relations among the variables. We use $PA_i$ to represent the parent set of $X_i$. The dependence between each variable $X_i$ and its parents is quantified using a conditional probability distribution, $P(X_i|PA_i)$. The product of the conditional probability distributions give the joint distribution over all of the variables.

We consider the score-based Bayesian network structure learning problem (BNSL) in this paper. Given a dataset $D = \{D_1, ..., D_N\}$, where $D_i$ is a complete instantiation of all of the variables $V$, and a scoring function $s$, the goal is to find a Bayesian network structure $S^*$ such that $S^* = \arg\min_S s(S, D)$. We omit $D$ for brevity in the remainder of the paper.

The scoring function is often a penalized log-likelihood or Bayesian criterion which trades off the goodness of fit of $S$ to $D$ against the complexity of $S$. We allow for any *decomposable* score, i.e., the score for $S$ is the sum of the scores of each variable, $s(S) = \sum_{X \in V} s(X, PA_X)$. Most commonly used scoring functions, including MDL (Lam and Bacchus 1994), fNML (Silander et al. 2008) and BDe (Buntine 1991; Heckerman 1996), are decomposable.

In our work, we adopt a shortest path perspective to the problem, so we assume the optimal structure minimizes $s$. Some scoring functions, such as BDe, assign high values to better networks. We can multiply all scores by $-1$ to convert the maximization into a minimization.

## 3 THE SHORTEST PATH PERSPECTIVE

Yuan *et al.* (2011) formulated BNSL as a shortest-path finding problem. Figure 1 shows an implicit search graph for four variables in which the shortest path search is performed. Each node in the graph corresponds to an optimal subnetwork over a unique subset of variables in the dataset. The *start* search node, at the top of the graph, corresponds to the empty variable set, while the bottom-most node with all variables is the *goal* node. Each edge in the search graph represents adding a new variable $X$ as a leaf to the optimal subnetwork over the existing variables, $U$. The new variable selects its optimal parents (according to the scoring function $s$) from $U$. The cost of the edge is equal to the score of the optimal parent set, which we denote $BestScore(X, U)$, i.e.,

$$cost(U \to U \cup \{X\}) = BestScore(X, U)$$
$$= \min_{PA_X \subseteq U} s(X, PA_X).$$

Based on this specification, a path from *start* to *goal* induces an ordering on the variables, based on the order in which they are added. Thus, we also call this graph the *order graph*. Each variable selects its optimal parents from variables which precede it in the ordering. Consequently, combining the parent set selections made on a path from *start* to *goal* gives the optimal network for that ordering, and the cost of that path corresponds to the score for that network. Therefore, the shortest path from *start* to *goal* corresponds to a globally optimal Bayesian network.

The computation of $BestScore(\cdot)$ is required for each edge visited during the search. Naively, this computation requires considering an exponential number of parent sets; however, several authors (Teyssier and Koller 2005; de Campos and Ji 2011) have noted that many parent sets are not optimal for any ordering of variables. Therefore, many local scores can be pruned before the search. Yuan and Malone (2012) developed a sparse data structure which takes advantage of this pruning to store only the *possibly optimal parent sets* (POPS) and to compute $BestScore(\cdot)$ with a linear number of bitwise operations.

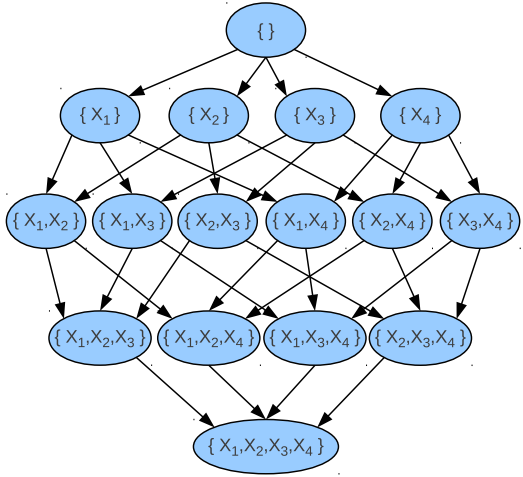

Figure 1: An order graph of four variables

This shortest path problem has been solved using several heuristic search algorithms, including A* (Yuan, Malone, and Wu 2011) and breadth-first branch and bound (BF-BnB) (Malone et al. 2011). In A* (Hart, Nilsson, and Raphael 1968), an admissible heuristic function is used to calculate a lower bound on the cost from a node $\mathbf{U}$ in the order graph to $goal$. An f-cost is calculated for $\mathbf{U}$ by summing the cost from $start$ to $\mathbf{U}$ (called $g(\mathbf{U})$) and the lower bound from $\mathbf{U}$ to $goal$ (called $h(\mathbf{U})$). So $f(\mathbf{U}) = g(\mathbf{U}) + h(\mathbf{U})$. The f-cost provides an optimistic estimation on how good a path can be if it has to go through $\mathbf{U}$. The search maintains a list of nodes to be expanded sorted by f-costs called $open$. It also keeps a list of nodes which have already been expanded called $closed$. Initially, $open$ contains just $start$, and $closed$ is empty. Nodes are then expanded in best-first order according to f-costs. Expanded nodes are added to $closed$. As better paths to nodes are discovered, they are added to $open$. In general, if a better path to a node in $closed$ is found, then the node must be added to $open$ again and re-expanded. Upon expanding $goal$, the shortest path from $start$ to $goal$ has been found.

In BFBnB, nodes are instead expanded one layer at a time, where a layer consists of all nodes corresponding to sub-networks of the same size. Before beginning the search, a local search strategy, such as greedy hill climbing, is used to find a "good" network and its score. During the BFBnB search, any node with an f-cost greater than the score found during the local search can safely be pruned.

Yuan *et al.* (Yuan, Malone, and Wu 2011) gave a simple heuristic function. Later, tighter heuristics based on pattern databases were developed (Yuan and Malone 2012).

All of the heuristics were shown to be *admissible*, i.e., to always give a lower bound on the cost from $\mathbf{U}$ to $goal$. Furthermore, the heuristics have been shown to be *consistent*, which is a property similar to non-negativity required by Dijkstra's algorithm. Primarily, in standard A*, consistency ensures that the first time a node is expanded, the shortest path to that node has been found, so no node ever needs to be re-expanded.

## 4 ANYTIME LEARNING ALGORITHMS

The shortest path perspective makes it straightforward to apply anytime heuristic search algorithms to solve the Bayesian network learning problem. The basic A* algorithm does not have anytime behavior. It expands nodes in best-first order until expanding $goal$ at which point it has the optimal solution. The heuristic search community has developed a variety of algorithms which allow A* to find solutions more quickly. We begin by discussing weighted A* (WA*), which does not add anytime behavior to A* but can greatly improve solving time while offering provable quality guarantees. We then discuss two techniques which add anytime behavior to WA*. We also describe a third strategy, not directly related to WA*, which adds anytime behavior and quality guarantees to A*.

### 4.1 WEIGHTED A*

Weighted A* (WA*) (Pohl 1970) is a variant of A* which adds a weight $\epsilon$ ($\geq 1.0$) to the heuristic function in the f-cost calculations. That is, $f(\mathbf{U}) = g(\mathbf{U}) + \epsilon \times h(\mathbf{U})$. Otherwise, WA* behaves exactly as unweighted A*. The inflated h-cost could now overestimate the cost of a path from $\mathbf{U}$ to the goal, so the calculation is no longer admissible or consistent. Despite the loss of admissibility, though, WA* still has a quality guarantee: if $h$ was originally consistent, the search algorithm can disregard any better paths it finds to nodes in $closed$, and the cost of the path found from $start$ to $goal$ is guaranteed to be no more than a factor of $\epsilon$ greater than the optimal solution.

Intuitively, much of the f-cost of nodes close to $start$ comes from $h$, but the f-cost of deeper nodes is dominated by $g$. Because WA* weights the $h$ costs, but not the $g$ costs, this has the effect of making the search favor deeper nodes because they have smaller $h$ values. Thus, the overall effect is that WA* expands deeper nodes more greedily than A* and often expands $goal$ much more quickly than the unweighted A*.

### 4.2 ANYTIME WEIGHTED A*

Like WA*, anytime weighted A* (AWeiA*) (Hansen and Zhou 2007) adds a weight to the heuristic calculations so it also favors expanding deeper nodes. Rather than stopping as soon as $goal$ is expanded, though, AWeiA* continues the search. During the search, a stream of better paths to *goal* are discovered, and the *incumbent* solution, which gives the current shortest path, is updated. If the search is run until completion, it terminates with the optimal solution. To guarantee the optimality of the final solution, though, AWeiA* must re-expand nodes when it finds a better path to them. Any node which has a worse unweighted f-cost than the incumbent can be pruned, though. At any time, the search can be stopped, and the incumbent solution returned. AWeiA* also offers the same guarantee of WA*: the globally optimal solution is guaranteed to be within a factor of $\epsilon$ of the incumbent. However, an even tighter error bound is available by calculating the ratio of the smallest unweighted f-cost of any open node and the incumbent.

### 4.3 ANYTIME REPAIRING A*

Anytime repairing A* (ARA*) (Likhachev, Gordon, and Thrun 2003) also starts as normal WA*. Upon finding a solution, ARA* decreases $\epsilon$ and searches again. At each iteration, the solution improves (or stays the same), so this algorithm also produces a stream of improved solutions. Additionally, because $\epsilon$ is decreased at each iteration, the quality guarantee tightens, as well. ARA* can also check the ratio between the smallest unweighted f-cost and the incumbent to look for an even better bound. The algorithm terminates with the optimal solution after completing an iteration in which $\epsilon = 1$.

Like AWeiA*, ARA* can also find a better path to a node during the search. Rather than immediately adding the node back to *open*, though, ARA* keeps these nodes in a separate list, *repair*. At the beginning of each iteration, rather than beginning the search at *start*, ARA* instead adds all of *repair* to *open*. In this manner, ARA* reuses g-cost information from one iteration to the next. ARA* can also prune nodes with a worse f-cost than the incumbent.

### 4.4 ANYTIME WINDOW A*

Unlike AWeiA* and ARA*, anytime window A* (AWinA*) (Aine, Chakrabarti, and Kumar 2007) is not based on WA*. Rather, it uses a type of sliding window to encourage deeper exploration of the order graph. It consists of a series of iterations in which a parameter $w$, which increases from one iteration to the next, controls the size of the window. The algorithm keeps track of the depth of all nodes expanded during an iteration of the algorithm. After expanding a node in layer $l$, all nodes in layer $l - w$ are *frozen* for that iteration. Frozen nodes are stored, but are not expanded. When $h$ is consistent (like the heuristic functions used here in BNSL), AWinA* will only expand a node at most once during each iteration. Therefore, node re-expansions are not explicitly considered in this algorithm. As with the other anytime algorithms, AWinA* can prune any node with a worse f-cost than the incumbent. Similar to ARA*, rather than starting each iteration at *start*, AWinA* begins by adding all frozen nodes from the previous iteration to *open*, so it also reuses information across iterations.

AWinA* uses the sliding window to encourage more greedy behavior in the search, but there is no quality guarantee for window size similar to that of the weighted algorithms. This algorithm can calculate the ratio between the smallest unweighted f-cost of a frozen node and the incumbent to find a quality guarantee, though.

## 5 EMPIRICAL EVALUATIONS

We empirically evaluated the anytime weighted A* (AWeiA*), anytime window A* (AWinA*), and anytime repairing A* (ARA*) against the integer linear programming algorithm (GOBNILP, v1.1) (Cussens 2011) on Bayesian network learning tasks. The A* implementations are available online (http://url.cs.qc.cuny.edu/software/URLearning.html). For AWeiA*, we used a weight of 1.25. For ARA*, we also set the initial weight to be 1.25 and decreased it by 0.05 at each iteration. The initial window size of AWinA* was 0 and increased by 1 after each iteration. We used static pattern databases as the heuristic function. We empirically determined these parameters give good performance on a variety of datasets. We used the default parameter setting for the GOBNILP algorithm (ILP for short). We did not compare to local search strategies, such as greedy hill climbing or optimal reinsertion, because a previous study (Malone and Yuan 2012) has shown that WA* outperforms those algorithms. That study also showed that WA* outperformed BB (de Campos and Ji 2011). The first iteration of ARA* is equivalent to WA*, so we assume the results of that study extend here.

One objective in this study is to compare the anytime behavior of these algorithms, including the quality (i.e, score) of the anytime solution and the error bounds. The other objective is better understand the shortest-path formulation. To rigorously study both objectives, we generated test datasets by sampling synthetic Bayesian networks. For all experiments, we first selected a number of variables and maximum number of parents allowed for each variable. We then created the networks using a slight variation on the Ide-Cozman MCMC algorithm (Ide and Cozman 2002). During the MCMC process, if a successor network resulted in a variable exceeding the maximum number of parents, that network was discarded, and the MCMC continued from the previous network. All variables were binary, and conditional probability distributions were sampled from a symmetric Dirichlet distribution with a concentration parameter of 1. We call these networks the *generating networks*. Then, we generated datasets with 1,000 to 20,000 data points with logic sampling.

| Dataset | POPS | Best score | 60 Second Score | | | 60 Second Error Bound | | | Final Score | | | Final Error Bound | | | Running Time (seconds) | | |
|---|---|---|---|---|---|---|---|---|---|---|---|---|---|---|---|---|---|
| | | | AWinA* | ILP ARA* | AWeiA* | AWinA* | ILP ARA* | AWeiA* | AWinA* | ILP ARA* | AWeiA* | AWinA* | ILP ARA* | AWeiA* | AWinA* | ILP ARA* | AWeiA* |
| 29.3.1k | 2164 | **15298.15** | **1.000 1.000** 1.004 1.011 | | | 1.033 **1.000** 1.050 1.079 | | | **1.000 1.000** 1.004 1.002 | | | 1.015 **1.000** 1.050 1.069 | | | 980* 45 998* 2140* | | |
| 29.3.5k | 20033 | 73927.94 | **1.000** 1.007 1.012 1.015 | | | 1.065 **1.025** 1.100 1.113 | | | **1.000 1.000** 1.004 1.006 | | | 1.033 **1.006** 1.050 1.102 | | | 1151* 7194 713* 219* | | |
| 29.3.10k | 47598 | 147625.66 | **1.001** 1.011 1.014 1.008 | | | 1.072 **1.058** 1.124 1.118 | | | **1.000** 1.003 1.008 1.008 | | | 1.039 **1.016** 1.100 1.118 | | | 1333* 7178 1587* 19* | | |
| 29.3.20k | 124564 | 293081.53 | **1.002** - 1.012 1.010 | | | **1.093** - 1.133 1.131 | | | **1.000** 1.002 1.007 1.010 | | | 1.044 **1.022** 1.100 1.131 | | | 7261 7184 456* 2* | | |
| 29.6.1k | 1568 | **15478.77** | **1.000 1.000** 1.003 1.015 | | | 1.014 **1.001** 1.041 1.060 | | | **1.000 1.000 1.000** 1.005 | | | **1.000** 1.001 **1.000** 1.050 | | | 279 6 290 181* | | |
| 29.6.5k | 16141 | **72054.89** | **1.000** 1.002 1.003 1.010 | | | 1.030 **1.020** 1.050 1.081 | | | **1.000 1.000 1.000** 1.001 | | | **1.000 1.000 1.000** 1.071 | | | 283 6394 296 451* | | |
| 29.6.10k | 51542 | **141306.05** | **1.000** 1.057 1.007 1.010 | | | **1.061** 1.169 1.086 1.100 | | | **1.000 1.000** 1.001 1.001 | | | **1.000 1.000 1.000** 1.090 | | | 871 7200 1031 674* | | |
| 29.6.20k | 175340 | **279387.44** | **1.000** - 1.009 1.006 | | | **1.077** - 1.100 1.115 | | | **1.000 1.000 1.000** 1.006 | | | 1.001 1.021 **1.000** 1.115 | | | 7208 7140 5127 59* | | |
| 31.3.1k | 1290 | **16566.76** | 1.001 **1.000** 1.009 1.010 | | | 1.053 **1.000** 1.082 1.083 | | | **1.000 1.000** 1.004 1.010 | | | 1.027 **1.000** 1.050 1.083 | | | 982* 5 1161* 26* | | |
| 31.3.5k | 7885 | 80348.9 | 1.002 1.006 1.020 1.015 | | | 1.096 **1.022** 1.136 1.129 | | | **1.000 1.000** 1.008 1.005 | | | 1.067 **1.004** 1.100 1.119 | | | 864* 7199 382* 463* | | |
| 31.3.10k | 20529 | 160304.6 | **1.001** 1.007 1.026 1.006 | | | 1.114 **1.034** 1.159 1.138 | | | **1.000** 1.003 1.016 1.006 | | | 1.084 **1.016** 1.141 1.138 | | | 938* 7181 1071* 44* | | |
| 31.3.20k | 52459 | 319897.03 | **1.001** 1.016 1.020 1.010 | | | **1.122** 1.139 1.162 1.151 | | | **1.000** 1.003 1.014 1.010 | | | 1.099 **1.025** 1.150 1.151 | | | 1086* 7167 380* 4* | | |
| 31.6.1k | 1891 | **18758.37** | 1.007 **1.000** 1.017 1.013 | | | 1.030 **1.000** 1.046 1.042 | | | **1.000 1.000** 1.017 1.003 | | | 1.002 **1.000** 1.038 1.032 | | | 1017* 23 877* 279* | | |
| 31.6.5k | 28299 | **83224.24** | **1.000** 1.045 **1.000** 1.001 | | | **1.000** 1.057 **1.000** 1.041 | | | **1.000 1.000 1.000** 1.001 | | | **1.000 1.000 1.000** 1.041 | | | 13 2922 14 14* | | |
| 31.6.10k | 98824 | **161836.84** | **1.000** - **1.000** 1.001 | | | **1.000** - 1.017 1.052 | | | **1.000 1.000** 1.001 1.001 | | | **1.000 1.000 1.000** 1.052 | | | 42 7185 63 13* | | |
| 31.6.20k | 375404 | **318807.50** | **1.000** - 1.016 1.003 | | | **1.041** - 1.048 1.066 | | | **1.000** 1.011 **1.000** 1.003 | | | **1.000 1.000 1.000** 1.066 | | | 277 7193 334 37* | | |
| 33.3.1k | 2075 | **17356.83** | 1.006 **1.000** 1.019 1.022 | | | 1.072 **1.000** 1.100 1.111 | | | 1.001 **1.000** 1.019 1.012 | | | 1.040 **1.000** 1.096 1.101 | | | 910* 24 881* 1226* | | |
| 33.3.5k | 18491 | 83689.7 | 1.006 1.008 1.030 1.020 | | | 1.121 **1.039** 1.166 1.155 | | | **1.000** 1.002 1.019 1.011 | | | 1.069 **1.013** 1.148 1.145 | | | 1111* 7182 70* 279* | | |
| 33.3.10k | 55083 | 167095.75 | 1.004 1.020 11.025 1.012 | | | **1.131** 1.158 1.173 1.159 | | | **1.000** 1.002 1.015 1.012 | | | 1.089 **1.022** 1.150 1.159 | | | 1308* 7126 4302* 23* | | |
| 33.3.20k | 184523 | 333144.78 | **1.010** - 1.015 1.014 | | | **1.163** - 1.179 1.178 | | | **1.000** 1.002 1.015 1.006 | | | 1.104 **1.030** 1.179 1.168 | | | 7221 6976 39* 1093* | | |
| 33.6.1k | 2179 | **19101.15** | 1.008 **1.000** 1.021 1.015 | | | 1.025 **1.000** 1.045 1.041 | | | **1.000 1.000** 1.021 1.005 | | | 1.001 **1.000** 1.034 1.031 | | | 1072* 49 949* 88* | | |
| 33.6.5k | 29393 | **86040.53** | **1.000** 1.032 1.001 1.015 | | | 1.001 1.038 1.020 1.058 | | | **1.000 1.000 1.000** 1.005 | | | **1.000 1.000 1.000** 1.048 | | | 66 4073 110 80* | | |
| 33.6.10k | 93186 | **168673.1** | **1.000** - 1.015 1.013 | | | **1.026** - 1.050 1.066 | | | **1.000 1.000** 1.001 1.003 | | | **1.000 1.000** 1.005 1.055 | | | 243 7173 255 114* | | |
| 33.6.20k | 357832 | 331711.44 | **1.000** - 1.006 1.015 | | | 1.050 - **1.046** 1.082 | | | **1.000 1.000** 1.012 1.001 | | | **1.000 1.000** 1.021 1.061 | | | 1397 6936 1425 2561* | | |
| 35.3.1k | 4321 | **16935.32** | 1.003 1.005 1.021 1.033 | | | 1.065 **1.010** 1.100 1.124 | | | **1.000 1.000** 1.021 1.024 | | | 1.036 **1.000** 1.092 1.114 | | | 938* 84 749* 890* | | |
| 35.3.5k | 46429 | 81636.25 | 1.005 1.026 1.022 1.018 | | | 1.120 **1.109** 1.142 1.154 | | | **1.000** 1.003 1.022 1.009 | | | 1.072 **1.019** 1.142 1.144 | | | 1307* 7123 24* 4234* | | |
| 35.3.10k | 151501 | 162883.4 | **1.008** - 1.039 1.021 | | | **1.145** - 1.197 1.176 | | | 1.003 **1.000** 1.020 1.012 | | | 1.097 **1.026** 1.150 1.166 | | | 7245 7021 2779* 237* | | |
| 35.3.20k | 462115 | 326013.53 | **1.010** - 1.022 1.017 | | | **1.176** - 1.187 1.191 | | | **1.000 1.000** 1.022 1.009 | | | 1.117 **1.039** 1.187 1.181 | | | 7232 7185 57* 112* | | |
| 35.6.1k | 2111 | **20536.60** | 1.019 **1.000** 1.029 1.018 | | | 1.052 **1.000** 1.070 1.058 | | | 1.008 **1.000** 1.029 1.018 | | | 1.029 **1.000** 1.066 1.058 | | | 928* 40 572* 48* | | |
| 35.6.5k | 31608 | **91850.43** | **1.000** 1.031 1.025 1.039 | | | **1.022** 1.039 1.068 1.097 | | | **1.000 1.000 1.000** 1.011 | | | **1.000 1.000 1.000** 1.067 | | | 799 6087 1089 848* | | |
| 35.6.10k | 108155 | **179521.58** | **1.000** - 1.018 1.032 | | | **1.050** - 1.060 1.100 | | | **1.000 1.000** 1.001 1.001 | | | **1.000** 1.010 **1.000** 1.070 | | | 2024 7199 4993 2788* | | |
| 35.6.20k | 401515 | 353827.4 | **1.005** - 1.100 1.056 | | | **1.080** - 1.190 1.140 | | | **1.000** 1.021 1.003 1.002 | | | **1.010** 1.040 1.050 1.070 | | | 7203 4229 685* 5359* | | |

Table 1: A comparison of the quality of 60-second and final solutions plus error bounds of the algorithms. A dataset named "a.b.ck" stands for one with $a$ variables, maximum $b$ parents, and $ck$ data points. "POPS" shows the number of possibly optimal parent sets. "Best score" shows the MDL score of the best solution found by any of the algorithms; a boldfaced number indicates it was proven optimal. "60 Second Score" and "60 Second Error Bound" are the score (shown as the ratio over the best score) and error bound of solutions found by each algorithm after 60 seconds. A boldfaced number here indicates the best result among all algorithms. An error bound of "1.000" means a solution is proven optimal. "-" means that ILP did not find any solution within 60 seconds. Similarly, "Final Score" and "Final Error Bound" are for the solutions found after 2 hours. Finally, the running time until the last error bound improvement of the algorithms are reported in the column "Running Time". The total runtimes are sometimes slightly longer than 2 hours because of discrepancies between wall time and CPU time. An asterisk in the running time means the algorithm exhausted RAM.

All of the algorithms we consider require a decomposable scoring function. In this evaluation, we use the MDL scoring function (Lam and Bacchus 1994). Let $r_i$ be the number of states of $X_i$, $N_{pa_i}$ be the number of data points consistent with $PA_i = pa_i$, and $N_{x_i,pa_i}$ be the data points further constraint with $X_i = x_i$. Then MDL is given as follows.

$$MDL(G) = \sum_i MDL(X_i, PA_i), \quad (1)$$

where

$$MDL(X_i, PA_i) = H(X_i, PA_i) + \frac{\log N}{2} K(X_i, PA_i),$$
$$H(X_i, PA_i) = -\sum_{x_i, pa_i} N_{x_i, pa_i} \log \frac{N_{x_i, pa_i}}{N_{pa_i}},$$
$$K(X_i, PA_i) = (r_i - 1) \prod_{X_l \in PA_i} r_l.$$

### 5.1 ANYTIME RESULTS

We first tested the anytime behavior of the algorithms on random networks with $\{29, 31, 33, 35\}$ variables and $\{3, 6\}$ maximum parents per variable. Then, from each network, we generated datasets with $\{1k, 5k, 10k, 20k\}$ data points. Thus, in total, we considered 32 datasets. We put a 2-hour (7200 seconds) time limit on all the algorithms. The algorithms may terminate earlier than the time limit when either a provably optimal solution is found or RAM is exhausted. All of the algorithms (shortest-path-based and ILP) require the local scores ($MDL(X_i, PA_i)$) as input; therefore, we do not include these calculation times in the results. Table 1 shows all the results. We focus on synthetic networks in this study because we can control the parameters of the generating network; however, results from real-world data show similar trends.

The results show that AWinA* performs much better than the other shortest-path-based algorithms. Its 60-second and final scores and error bounds are better than those of AWeiA* and ARA* on almost all cases. We note, however, that AWinA* often runs longer than the other algorithms before exhausting all the RAM. This is because AWinA* finds better solutions more quickly than AWeiA* and ARA*. Therefore, it prunes more nodes during the search and explores more of the search space. Consequently, it fills RAM more slowly and is able to run longer. For these reasons, we only consider AWinA* among the shortest-path-based algorithms for the remaining discussion.

ILP performed quite well on all the datasets with only 1k data points; it found all the optimal solutions within 60 seconds. AWinA* also often found the optimal solutions quickly, although it took much longer in proving the optimality, and sometimes failed to do so before running out of memory. The reason for ILP's excellent performance is that the numbers of possibly optimal parent sets (POPS) for these datasets are quite small. Therefore, the linear programs constructed by ILP are small and easy to solve.

However, AWinA*'s 60-second and final solutions are all better than those found by ILP on the datasets with 5k, 10k and 20k data points, even though the error bound is sometimes worse than that of ILP. Those datasets had many more POPS, so each iteration of ILP required solving a large linear program. As a result, ILP was sometimes not able to find any solution within 60 seconds. The difference between scores at 60-seconds and the end of the search also show that ILP does not typically find its best solution early in the search. In contrast, AWinA* was always able to find a solution within several seconds. In fact, AWinA* found its best solution within the first 60 seconds on 14 of the datasets; the rest of the time is spent on proving the optimality of the solutions. This behavior is highly desirable. For a given large dataset, we do not know whether an optimal Bayesian network can be learned given limited resources. We should therefore strive to obtain as good a solution as we can as quickly as possible. The results show that AWinA* finds better solutions much sooner than ILP on many of the test datasets.

We note, however, ILP sometimes provides better error bounds than AWinA*. This is surprising given that its solutions are of lower quality, in terms of score, than those of AWinA*. The reason is that ILP often finds tighter lower bounds than AWinA*. The solutions of the LPs for ILP optimize the relaxed ILPs, so they give a lower bound on the solution. A simple local search is used to extract a valid BN from the LP solution and attempt to improve the upper bound. So most of the work in ILP focuses on improving the lower bound. On the other hand, the primary goal of AWinA* is to find a shortest path (subject to the sliding window constraint) from $start$ to $goal$, which improves the upper bound. Shortest-path-based algorithms must expand nodes with the lowest f-costs to improve the lower bounds, but the window semantics (and also weighted heuristic) discourages expanding nodes in early layers of the search, regardless of their f-cost. Therefore, AWinA* focuses more heavily on improving the upper bound.

The better error bounds are certainly nice to have. AWinA* proved the optimality of its solutions for 13 of the datasets, while ILP proved optimality for 14. However, based on the error bounds of ILP, we can verify that AWinA* actually found optimal solutions for several other datasets. For example, for the "29.3.1k" dataset, both algorithms found a network with a score of $15,298.15$, but the final error bound for ILP is $1.00$, while the bound for AWinA* is $1.07$. Given ILP's error bound, then, we can conclude that AWinA* actually found the optimal network. Using this line of reasoning, we can see that AWinA* found the optimal network on 16 datasets, but ILP on only 14. The results suggest that ILP always either found-and-proved or did-not-find the optimal network on these datasets.

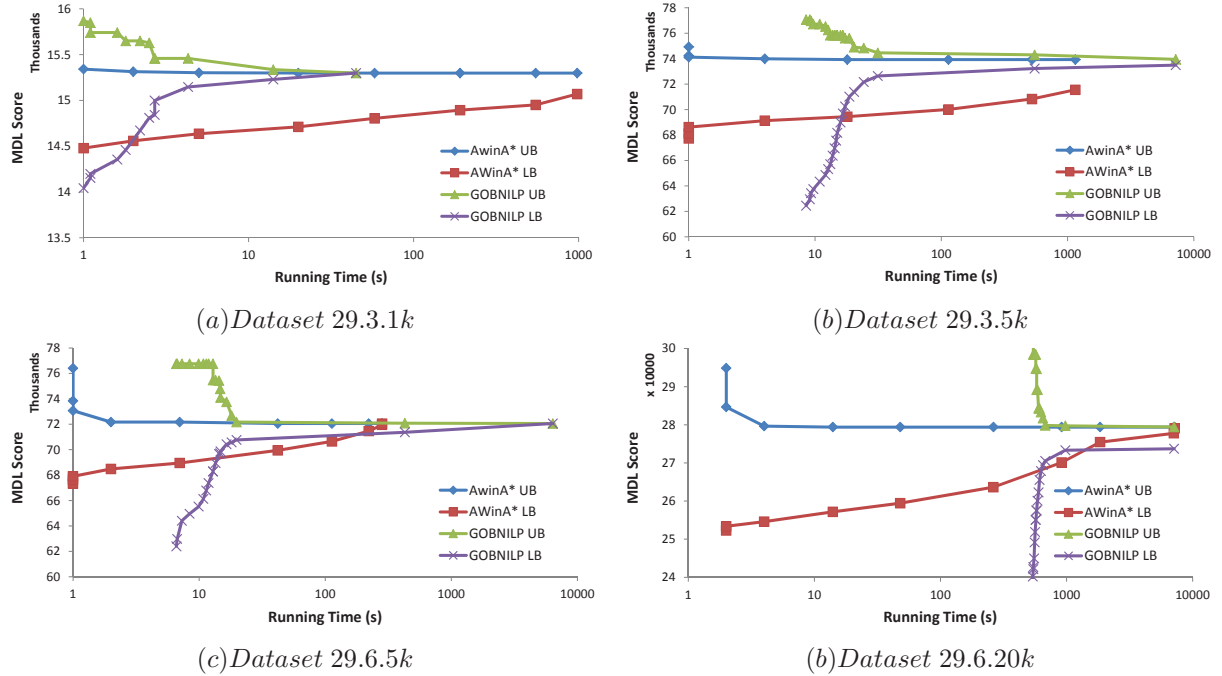

Figure 2: A comparison of the convergence of upper bounds (UB) and lower bounds (LB) for AWinA* and ILP.

Another difference between the two algorithms is that AWinA* often terminates before the time limit because it exhausts all of the available RAM storing *open* and *closed* lists. On the other hand, ILP typically runs out of time, but does not fully utilize RAM. These results suggest that ILP may be able to find solutions of the same quality as AWinA* if given enough time, but it is unclear how much more time would be needed. Similarly, more RAM or an external-memory strategy could improve the quality and error bounds of the solutions of AWinA*.

## 5.2 CONVERGENCE OF BOUNDS

To gain a better perspective on how AWinA* and ILP improve the upper and lower bounds, we plot the convergence curves of the bounds against the running time for several datasets in Figure 2. The results clearly agree with our analysis in Section 5.1. AWinA* was able to find good solutions very quickly, while ILP was slower to find its first solution. Also, even though ILP finds quite bad lower bounds initially, it quickly improves them. Finally, the pace with which AWinA* improves its solutions and error bounds is quite regular (increasing roughly exponentially from one iteration to the next). In comparison, ILP was able to improve its solution quickly and often in the early stage of the search, but its pace slowed down significantly in the later stages. This suggests that ILP may need much longer to find the next solution. For the "29.3.1k" dataset, ILP found and proved the optimal network in 45 seconds. Even though AWinA* initially found better solutions than ILP, and the optimal solution in 58 seconds, it failed to prove its optimality before running out of memory. For the "29.3.5k" dataset, both algorithms failed to prove the optimality of their solutions. AWinA* found its best solution in 18 seconds. Based on its behavior on other datasets, we suspect that AWinA* found the optimal solution but ran out of RAM before proving its optimality. ILP's solution was worse than that of AWinA*, so it definitely did not find the optimal solution; however, it obtained a much better lower bound and, hence, a better error bound. For the "29.6.5k" dataset, both algorithms were able to prove optimality of the solutions. AWinA* was able to find the optimal solution in 42 seconds and prove it in 283 seconds, while ILP only finds the optimal solution near the end of the search (6,394s). Finally for the "29.6.20k" dataset, AWinA* found the optimal solution in 14 seconds and proved its optimality close to the time limit. ILP took much longer (497s) before finding its first solution, and was not able to find the optimal solution within the time limit.

## 5.3 EFFECT OF GENERATING PARAMETERS

We created the generating networks by varying the numbers of variables and maximum parents allowed for each variable, and generated the testing datasets with different numbers of data points. In general, more variables or more data points makes a dataset more difficult to solve optimally, and, hence, increased the error bounds of both AWinA* and ILP. Relatively, the number of data points has a larger effect on ILP; it solved almost all 1k datasets and several 5k datasets optimally, but none of the 10k or 20k datasets. The reason is that more data points increase the number of POPS and make the linear programs larger and harder to solve.

| Dataset | Percentage | Generating | Learned | Distance |
|---|---|---|---|---|
| 29.2.1k | 41.92 | 1.72 | 1.62 | 7 |
| 29.2.5k | 80.16 | 1.72 | 1.69 | 3 |
| 29.2.10k | 89.40 | 1.72 | 1.72 | 0 |
| 29.2.20k | 94.37 | 1.72 | 1.72 | 0 |
| 29.4.1k | 1.79 | 3.03 | 3 | 1 |
| 29.4.5k | 8.81 | 3.03 | 3.03 | 0 |
| 29.4.10k | 18.54 | 3.03 | 3.03 | 0 |
| 29.4.20k | 34.35 | 3.03 | 3.03 | 0 |
| 29.6.1k | 0.22 | 4.24 | 3.45 | 23 |
| 29.6.5k | 0.03 | 4.24 | 4.24 | 3 |
| 29.6.10k | 0.11 | 4.24 | 4.21 | 1 |
| 29.6.20k | 0.44 | 4.24 | 4.21 | 1 |
| 29.8.1k | 18.59 | 5.03 | 2.52 | 95 |
| 29.8.5k | 0.48 | 5.03 | 4.66 | 11 |
| 29.8.10k | 0.23 | 5.03 | 4.97 | 9 |
| 29.8.20k | 0.20 | 5.03 | 5.03 | 0 |

Table 2: The percentage of search nodes with better f-cost than the optimal solution ("Percent"); the average number of parents in the original ("Original") and learned networks ("Learned"), and the structural Hamming distance between the original and learned networks ("Distance").

A somewhat surprising observation comes from the effect of the maximum allowed parents in the generating networks on the the algorithms. AWinA* was able to solve almost all the datasets that allow up to 6 parents (6-parent datasets for short hereafter) optimally, but none of the 3-parent datasets. Similarly, ILP was able to find optimal solutions for more 6-parent datasets than 3-parent datasets. This is surprising because, intuitively, more parents make the Bayesian networks more complex and seemingly harder to learn. To better understand the effect of the generating parameters on the algorithms, especially AWinA*, we performed a more detailed sensitivity analysis.

### 5.4 SENSITIVITY ANALYSIS OF PARAMETERS

To study the sensitivity of shortest-path-based algorithms to the parameters for network and dataset generation, we generated networks with: 29 variables and {2, 4, 6, 8} maximum parents per variable. Then, from each network, we again generated datasets with: {1k, 5k, 10k, 20k} data points. We take the number of parameters necessary to specify the conditional probability distributions in the generating network as a measure of complexity (i.e., more parameters mean a more complex process).

For each dataset, we first collected the f-costs of all the nodes in the order graph for each dataset using a BFS search. Because we were interested only in the characteristics of the search space, we did not impose any time limit on the algorithm; it can effectively use external memory, so that resource did not pose a problem, either. Table 2 shows the percentages of search nodes that have better f-costs than the optimal solution, as well as several other statistics. We also show distributions of the f-costs of all nodes relative to the optimal solutions in Figure 3.

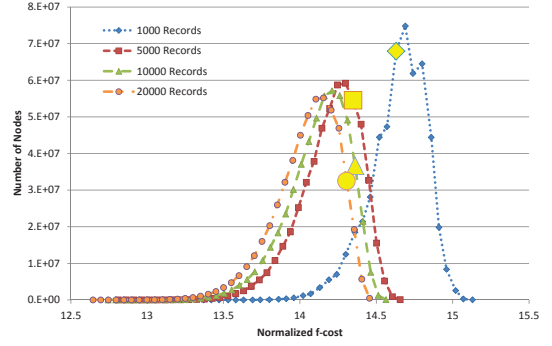

(a) Datasets 29.2.*

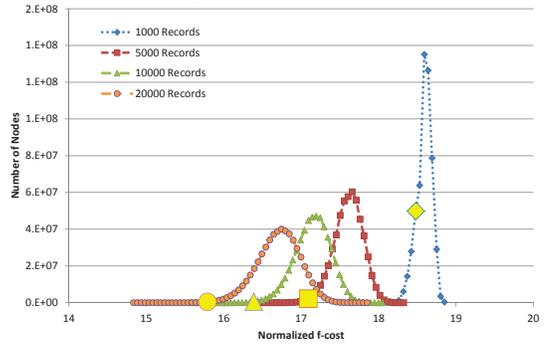

(b) Datasets 29.8.*

Figure 3: Distributions of the f-costs of all the search nodes normalized by the number of data points. The enlarged marks indicate where the optimal solutions are located.

As shown in Equation 1, the MDL score consists of two terms: the log-likelihood of the data given the structure and a structure complexity penalty. The likelihood term increases linearly in the number of data points, while the term that penalizes structural complexity increases logrithmically in the number of data points. Figure 4 shows that when the number of data points is small, possibly optimal parent sets (POPS) are typically smaller; consequently, learned optimal networks tend to be simpler than generating networks. As the number of data points increases, POPS become larger and learned networks more complex.

Indeed, as Table 2 shows, the average number of parents in the learned, optimal network increases as the number of data points increases (except for a slight decrease from 5k to 10k data points for the 6-parent networks). As the number of data points increases, the structural Hamming distance between the learned, optimal network and the generating network decreases and drops to 0 for several datasets. This shows that, given enough data, MDL can recover the generating network and is appropriate for study. In addition, Figure 3 shows that the normalized f-costs of the optimal solutions shift left with increasing data points; this is because more variables used larger parent sets and obtained better scores. We also observe that the f-cost distributions of all search nodes shifted leftward. Because of the heuristic functions used in A*, the internal order graph nodes relax the acyclic constraints between some of the variables, and have even more freedom to use the larger POPS. As a result, more internal nodes obtained better f-costs.

Therefore, the percentages of f-costs better than the optimal solutions depend on the relative speed in which the optimal solutions and f-cost distributions shift. For the 8-parent datasets, when we have few data points, even though the generating network is complex, the relatively large complexity penalty forces the search to consider simple networks which do not explain the data as well but incur a small complexity penalty. As Table 2 shows, based on the average number of parents in the generating network compared to the optimal network for 1k records, the optimal network is quite a bit simpler than the generating network. As we add more data points, though, the likelihood term dominates the score calculations. Therefore, the complex structures which better explain the data have, relatively, much better scores than simpler structures. Figure 3(b) shows that the optimal solutions shift to the left relative to the other nodes for the 8-parent datasets as the number of data points increases. That explains why the percentage of nodes with better f-costs than the optimal network is high, but decreases as the number of data points increases.

It is a different story for the 2- and 4-parent datasets. As Table 2 shows, for the 2-parent datasets, the percentages of nodes with f-costs better than the optimal network are rather high. The percentages increase with the number of data points and approach 95% for the 20k dataset. For the 4-parent datasets, the percentage is initially low but increases significantly as the number of data points increases. To understand why, we again consult Table 2, which shows that the generating networks for those datasets do not have many parents for each node. Therefore, simpler structures both explain the data well and have a low complexity penalty. Unlike in the 8-parent case, more complex structures can not improve upon the likelihood very much but incur a much larger complexity penalty. Consequently, fewer data points are needed to predict the structures well. The results show that even with only 1k data points, the learned networks have similar numbers of parents and structures as the generating networks. So the learned, optimal networks have converged to the generating networks and do not benefit much from more data points. Figure 3(a) indeed shows that the optimal solutions did not shift left much with more than 5k data points; they actually shift towards the right tails of the distributions with more data points.

These results help explain the performance, particularly the error bounds, of AWinA* in the first set of experiments. To completely prove optimality, AWinA* would have to expand all nodes with better f-costs than the goal. In practice, though, it can only expand about 10 million nodes in a search space in the allocated resources. These results suggest that AWinA* would not be able to prove optimality for datasets generated from simple networks and a large number of data points. Table 1 shows this is exactly the case.

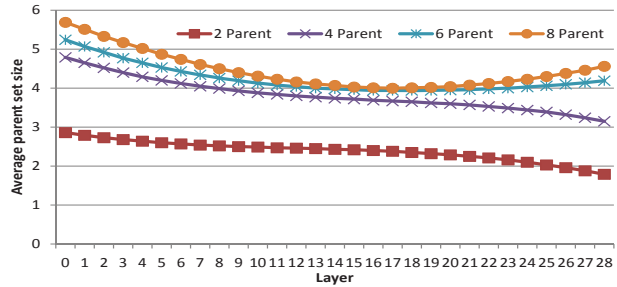

Figure 4: Average parent set size of all the nodes in each layer of the order graph for the "29.*.5k" datasets.

As further evidence, Figure 4 shows the average parent set size of the (cyclic) networks corresponding to all search nodes in each layer of the order graph for the "29.*.5k" datasets. For the 2- and 4-parent datasets, the average cardinality decreases monotonically. The heuristic estimates of most search nodes seem to select larger parent sets and have lower costs than the goal, so they would have to be expanded by A*. For the 6- and 8-parent datasets, the average cardinality dips initially and then increases. Many nodes in the middle layers seem to select smaller parent sets and have higher costs than the goal, so many of them are never expanded. The rate of the change of average cardinality is often larger in the beginning and last layers. The explanation is that the beginning and last layers have much fewer search nodes than the middle layers, so the changes in parent sets have a larger effect on the average cardinality.

## 6 CONCLUSIONS

In this research, we adapted several anytime heuristic search algorithms to learn optimal Bayesian networks from data, and empirically evaluated these algorithms against an integer linear programming algorithm. Our empirical results show that AWinA* is the best-performing anytime algorithm among the ones we evaluated in this study. It finds better, often optimal, solutions more quickly than existing methods; in many cases, the majority of its running time is spent on proving the optimality of a solution found early in the search. In comparison, the ILP algorithm focuses on finding lower bounds for the optimal solution in its search. As a result, its lower bounds are often better than those of AWinA*, even though its solutions are often not as good.

Our results show that, surprisingly, complex generating networks may seem structurally challenging to learn, but they actually lie within the promising solution space that is first explored by heuristic search and are easy to find.

**Acknowledgements** This work was supported by NSF grants IIS-0953723 and IIS-1219114.

# References

Aine, S.; Chakrabarti, P. P.; and Kumar, R. 2007. AWA*- a window constrained anytime heuristic search algorithm. In *Proceedings of the 20th international joint conference on Artifical intelligence*, IJCAI'07, 2250–2255. San Francisco, CA, USA: Morgan Kaufmann Publishers Inc.

Buntine, W. 1991. Theory refinement on Bayesian networks. In *Proceedings of the seventh conference (1991) on Uncertainty in artificial intelligence*, 52–60. San Francisco, CA, USA: Morgan Kaufmann Publishers Inc.

Chickering, D. M. 1996. Learning Bayesian networks is NP-complete. In *Learning from Data: Artificial Intelligence and Statistics V*, 121–130. Springer-Verlag.

Chickering, D. M. 2002. Learning equivalence classes of Bayesian-network structures. *J. Mach. Learn. Res.* 2.

Cussens, J. 2011. Bayesian network learning with cutting planes. In *Proceedings of the Twenty-Seventh Conference on Uncertainty in Artificial Intelligence (UAI-11)*, 153–160. Corvallis, Oregon: AUAI Press.

de Campos, C. P., and Ji, Q. 2011. Efficient learning of Bayesian networks using constraints. *Journal of Machine Learning Research* 12:663–689.

Hansen, E. A., and Zhou, R. 2007. Anytime heuristic search. *Journal of Artificial Intelligence Research* 28.

Hart, P. E.; Nilsson, N. J.; and Raphael, B. 1968. A formal basis for the heuristic determination of minimum cost paths. *IEEE Transactions On Systems Science And Cybernetics* 4(2):100–107.

Heckerman, D. 1996. A tutorial on learning with Bayesian networks. Technical report, Learning in Graphical Models.

Ide, J. S., and Cozman, F. G. 2002. Random generation of bayesian networks. In *Brazillian Symposium on Artificial Intelligence*, 366–375.

Jaakkola, T.; Sontag, D.; Globerson, A.; and Meila, M. 2010. Learning Bayesian network structure using LP relaxations. In *Proceedings of the 13th International Conference on Artificial Intelligence and Statistics (AISTATS)*.

Koivisto, M., and Sood, K. 2004. Exact Bayesian structure discovery in Bayesian networks. *Journal of Machine Learning Research* 549–573.

Lam, W., and Bacchus, F. 1994. Learning Bayesian belief networks: An approach based on the MDL principle. *Computational Intelligence* 10:269–293.

Likhachev, M.; Gordon, G.; and Thrun, S. 2003. ARA*: Anytime A* search with provable bounds on sub-optimality. In Thrun, S.; Saul, L.; and Schölkopf, B., eds., *Proceedings of Conference on Neural Information Processing Systems (NIPS)*. MIT Press.

Malone, B., and Yuan, C. 2012. A parallel, anytime, bounded error algorithm for exact bayesian network structure learning. In *Proceedings of the Sixth European Workshop on Probabilistic Graphical Models (PGM-12)*.

Malone, B.; Yuan, C.; Hansen, E.; and Bridges, S. 2011. Improving the scalability of optimal Bayesian network learning with external-memory frontier breadth-first branch and bound search. In *Conference on Uncertainty in Artificial Intelligence (UAI-11)*, 479–488. Corvallis, Oregon: AUAI Press.

Malone, B.; Yuan, C.; and Hansen, E. 2011. Memory-efficient dynamic programming for learning optimal Bayesian networks. In *National conference on Artifical intelligence*.

Moore, A., and Wong, W.-K. 2003. Optimal reinsertion: A new search operator for accelerated and more accurate Bayesian network structure learning. In *Intl. Conf. on Machine Learning*, 552–559.

Ott, S.; Imoto, S.; and Miyano, S. 2004. Finding optimal models for small gene networks. In *Pac. Symp. Biocomput*, 557–567.

Pohl, I. 1970. Heuristic search viewed as path finding in a graph. *Artificial Intelligence* 1(3-4):193 – 204.

Silander, T., and Myllymaki, P. 2006. A simple approach for finding the globally optimal Bayesian network structure. In *Proceedings of the 22nd Annual Conference on Uncertainty in Artificial Intelligence (UAI-06)*. Arlington, Virginia: AUAI Press.

Silander, T.; Roos, T.; Kontkanen, P.; and Myllymaki, P. 2008. Factorized normalized maximum likelihood criterion for learning Bayesian network structures. In *Proceedings of the 4th European Workshop on Probabilistic Graphical Models (PGM-08)*, 257–272.

Singh, A., and Moore, A. 2005. Finding optimal Bayesian networks by dynamic programming. Technical report, Carnegie Mellon University.

Teyssier, M., and Koller, D. 2005. Ordering-based search: A simple and effective algorithm for learning Bayesian networks. In *Proceedings of the Twenty-First Conference Annual Conference on Uncertainty in Artificial Intelligence (UAI-05)*, 584–590. Arlington, Virginia: AUAI Press.

Yuan, C., and Malone, B. 2012. An improved admissible heuristic for finding optimal bayesian networks. In *Proceedings of the Twenty-Eighth Conference on Uncertainty in Artificial Intelligence (UAI-12)*. AUAI Press.

Yuan, C.; Malone, B.; and Wu, X. 2011. Learning optimal Bayesian networks using A* search. In *Proceedings of the 22nd International Joint Conference on Artificial Intelligence*.